\documentclass[conference]{IEEEtran}
\usepackage{times}

\usepackage[numbers]{natbib}
\usepackage{multicol}
\usepackage[bookmarks=true]{hyperref}
\usepackage{url}
\usepackage{booktabs}
\usepackage{multirow}
\usepackage{graphicx}
\usepackage{xspace}
\usepackage{amsmath}
\usepackage{amssymb}
\usepackage{cuted}
\usepackage{subcaption}
\usepackage{xcolor}
\usepackage[table]{xcolor}
\usepackage{wrapfig}
\usepackage{cite}

\begin{document}

\definecolor{mypink}{RGB}{255,146,128}
\definecolor{mygreen}{RGB}{96,208,171}
\definecolor{mypink_light}{RGB}{255,146,128}
\definecolor{mygreen_light}{RGB}{216,242,232}
\definecolor{mypink_dark}{RGB}{250,90,80}
\definecolor{mygreen_dark}{RGB}{80,180,155}
\definecolor{DeepPink}{RGB}{255,20,147}

\newcommand{\red}[1]{\textcolor{red}{#1}}
\newcommand{\sizhe}[1]{\red{(Sizhe: {#1})}}
\newcommand{\lnxu}[1]{\textbf{\color{purple}#1}}
\newcommand{\mjc}[1]{\textbf{\color{orange}#1}}
\newcommand{\TODO}[1]{\textbf{\color{red}[TODO: #1]}}
\newcommand{\ours}{{Robo3R}\xspace}

\title{Robo3R: Enhancing Robotic Manipulation with Accurate Feed-Forward 3D Reconstruction}


\author{Sizhe Yang$^{1,2}$\quad Linning Xu$^{1,2,\dagger}$\quad Hao Li$^{1,3}$\quad Juncheng Mu$^{1,4}$\quad Jia Zeng$^{1}$\quad Dahua Lin$^{1,2}$\quad Jiangmiao Pang$^{1,\dagger}$ \\ 
$^{1}$Shanghai AI Laboratory 
\quad $^{2}$The Chinese University of Hong Kong \\
$^3$University of Science and Technology of China \quad $^4$Tsinghua University\\
\textcolor{gray}{$^{\dagger}$ Corresponding authors}\\
Project page: \textcolor{mypink_dark}{\url{https://yangsizhe.github.io/robo3r/}
}
}



%

\maketitle

\begin{strip}
\centering
\vspace{-8mm}
\includegraphics[width=\textwidth]{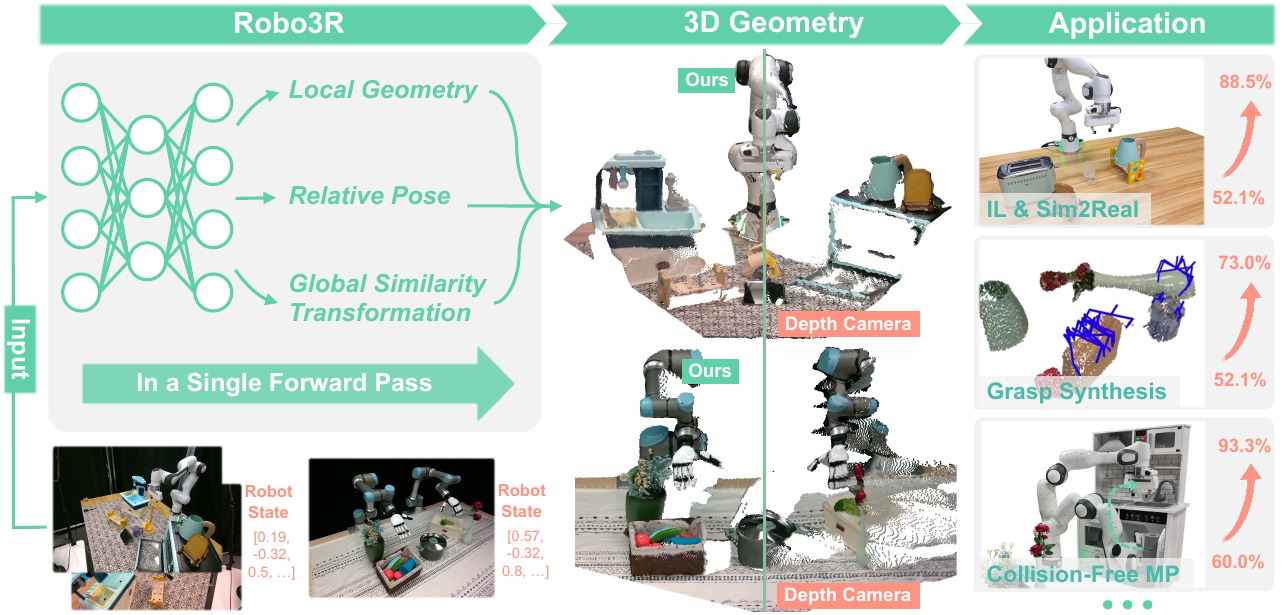}
\vspace{-3mm}
\captionof{figure}{\textbf{Overview.} 
\textbf{\ours} enables manipulation-ready 3D reconstruction from RGB frames in real time.
By achieving accurate metric-scale 3D geometry in the canonical robot frame, 
\ours \textbf{eliminates the need for depth sensors and calibration, while improving accuracy and robustness} in challenging manipulation scenarios.
These features lead to notable improvements in downstream applications such as imitation learning, sim-to-real transfer, grasp synthesis, and collision-free motion planning.}
\label{fig:teaser}
\end{strip}

\begin{abstract}
3D spatial perception is fundamental to generalizable robotic manipulation, yet obtaining reliable, high-quality 3D geometry remains challenging. 
Depth sensors suffer from noise and material sensitivity, while existing reconstruction models lack the precision and metric consistency required for physical interaction.
We introduce Robo3R, a feed-forward, manipulation-ready 3D reconstruction model that predicts accurate, metric-scale scene geometry directly from RGB images and robot states in real time. Robo3R jointly infers scale-invariant local geometry and relative camera poses, which are unified into the scene representation in the canonical robot frame via a learned global similarity transformation.
To meet the precision demands of manipulation, Robo3R employs a masked point head for sharp, fine-grained point clouds, and a keypoint-based Perspective-n-Point (PnP) formulation to refine camera extrinsics and global alignment.
Trained on Robo3R-4M, a curated large-scale synthetic dataset with four million high-fidelity annotated frames, Robo3R consistently outperforms state-of-the-art reconstruction methods and depth sensors.
Across downstream tasks including imitation learning, sim-to-real transfer, grasp synthesis, and collision-free motion planning, we observe consistent gains in performance, suggesting the promise of this alternative 3D sensing module for robotic manipulation.
\end{abstract}


\IEEEpeerreviewmaketitle

\section{Introduction}
\label{sec:introduction}

In unstructured environments, generalizable robotic manipulation relies heavily on robust spatial perception of the physical world. The perception of the external environment for robotic systems is generally categorized by input modality into 2D-based and 3D-based types. 
While RGB images are readily accessible, 2D-based policies often require extensive training data and frequently struggle with complex tasks~\citep{dp3}. In contrast, 3D-based policies are more data-efficient and generalizable.
Furthermore, for geometry-centric capabilities, such as grasp synthesis~\citep{anygrasp, dexgraspnet2} and collision-free motion planning~\citep{neuralmp, drp, curobo}, explicit 3D input is typically indispensable for ensuring accurate physical grounding.

However, acquiring 3D modalities is significantly more challenging than acquiring RGB images. Currently, 3D data for robotic manipulation is primarily obtained using depth cameras, which often produce depth maps of limited quality. Common depth cameras, whether stereo-based (e.g., RealSense D455) or time-of-flight based (e.g., Azure Kinect), are susceptible to noise and inaccuracies. Moreover, the quality of depth maps degrades drastically when encountering transparent or reflective objects, or under adverse lighting conditions.

Recently, the fields of depth estimation~\citep{da2, moge, moge2, pixel_perfect_depth, depth_pro} and feed-forward 3D reconstruction~\citep{dust3r, vggt,  mapanything, foundationstereo} have witnessed rapid advancement. 
However, despite their promise for enhancing spatial perception in robotics, most existing feed-forward reconstruction models still lack manipulation-level geometric precision and reliable metric scale, limiting their applicability in real-world robotic manipulation.

To address these challenges, we propose \textbf{\ours}, a feed-forward 3D reconstruction model designed specifically for robotic manipulation, as shown in Fig.~\ref{fig:teaser}. 
\ours fuses RGB images with robot states and processes the fused features using the Alternating-Attention mechanism, which facilitates efficient information propagation within and across frames.
\ours incorporates several specialized decoding heads. The proposed masked point head decomposes dense point prediction into depth, normalized image coordinate, and mask prediction, mitigating over-smoothing and producing sharp, fine-grained geometric details via unprojection and masking. The relative pose head predicts relative poses to register points across multiple views, while the similarity transformation (S.T.) tokens extract global similarity transformations, mapping points into metric-scale 3D geometry within the canonical robot frame.
An extrinsic estimation module extracts robot keypoints and estimates camera extrinsics by solving the Perspective-n-Point (PnP) problem~\citep{pnp}, further refining the similarity transformation. 
Compared to previous feed-forward reconstruction models, \ours explicitly incorporates robot priors, significantly enhancing reconstruction fidelity.
We train \ours on our curated large-scale high-fidelity synthetic dataset that contains four million frames. 
Thanks to the dataset's high diversity and photorealism, \ours successfully transfers to real-world manipulation scenarios.

We validate \ours through extensive experiments, demonstrating that it outperforms state-of-the-art feed-forward reconstruction models and depth sensors in terms of 3D geometry quality. The accurate 3D geometry produced by \ours enables improved performance in downstream applications, including real-world imitation learning, sim-to-real transfer, grasp synthesis, and collision-free motion planning. Crucially, our results highlight \ours's superior robustness in challenging scenarios: it effectively handles transparent, reflective, and tiny objects that typically hinder depth cameras.

In summary, the contributions of this paper are threefold:

\begin{enumerate}
    \renewcommand{\labelenumi}{\arabic{enumi})}

    \item We introduce \textbf{Robo3R-4M}, a large-scale synthetic dataset and corresponding data pipeline designed for perception and reconstruction in robotic manipulation scenarios. The dataset comprises four million frames and features diverse assets, extensive randomization, rich modalities, and annotations.

    \item We propose \textbf{\ours}, a feed-forward 3D reconstruction model tailored for robotic manipulation, which achieves high-fidelity depth estimation, precise camera parameter prediction, accurate metric scaling, and maintains a canonical coordinate system in real time.

    \item We conduct extensive qualitative and quantitative experiments validating \ours as a superior alternative to depth cameras. The results demonstrate that \ours produces higher-quality 3D representation, exhibits greater robustness to varying object materials and challenging scenarios, and significantly enhances spatial perception for robotic manipulation.
\end{enumerate}

\begin{figure*}[h!]
    \centering
    \includegraphics[width=0.98\linewidth]{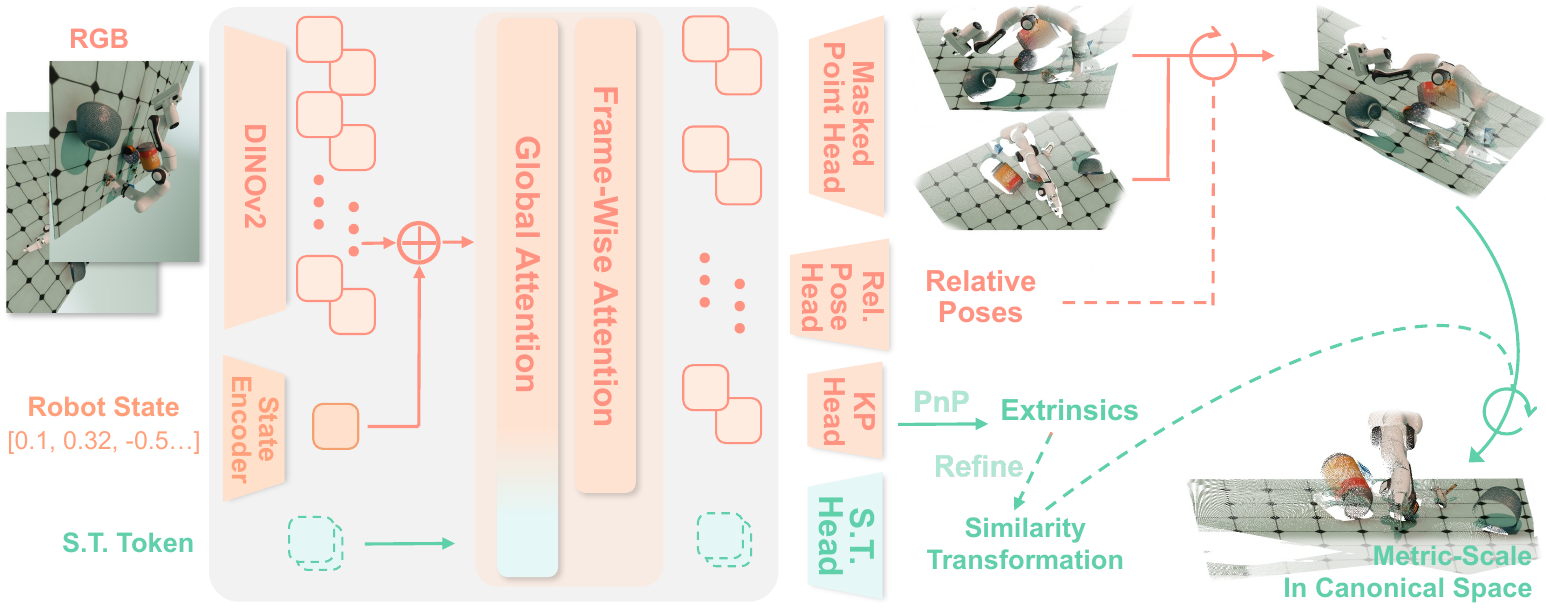}
    \caption{\textbf{Method Overview}. RGB images and robot states are encoded and fused. The transformer backbone processes the resulting features through alternating global and frame-wise attention. The masked point head decodes scale-invariant local geometry, while the relative pose head outputs relative poses for registering points across multiple views. S.T. tokens read out the global similarity transformation, which maps the points into metric-scale 3D geometry in the canonical robot frame. 
    }
    \label{fig:model}
    \vspace{-3mm}
\end{figure*}

\section{Related Work}
\label{sec:related_work}

\noindent\textbf{Feed-forward 3D reconstruction.}
In recent years, 3D reconstruction has rapidly advanced, particularly with the adoption of feed-forward neural networks.
DUSt3R~\citep{dust3r} has first shown promising results in this direction, predicting point maps from two images. VGGT~\citep{vggt} takes a further step by enabling 3D reconstruction from one, a few, or even hundreds of input views.
Several studies address the reconstruction of dynamic scenes from videos~\citep{monst3r, wang2025continuous, d4rt, trace_anything, sucar2025dynamic, feng2025st4rtrack}.
$\pi^3$~\citep{pi3} utilizes a permutation-equivariant architecture to estimate affine-invariant camera poses and scale-invariant local point maps. MapAnything~\citep{mapanything} supports optional geometric inputs and achieves metric scale reconstruction. DepthAnything3~\citep{da3} further simplifies feed-forward 3D reconstruction by using a single plain transformer and a minimal set of depth-ray prediction targets.
Nevertheless, these approaches mainly focus on scene-level reconstruction and exhibit limited accuracy in fine-grained geometry and scale, which are crucial for robotic manipulation.
Our work addresses these limitations through the synthesis of high-quality data and careful model design.

\noindent\textbf{3D representation for manipulation.}
Compared to RGB images, 3D representation provides better physical grounding and are therefore widely used in robotic manipulation. 
Several studies have employed 3D representation as model inputs to enable efficient and generalizable policy learning~\citep{dp3, maniflow, act3d, peract, gnfactor, chaineddiffuser, 3d_diffuser_actor, rvt, rvt2, dnact, h3dp, opfa}.
It has also been shown that 3D representation is more suitable than RGB images for sim-to-real transfer of manipulation policies~\citep{dextrah, maniwhere, dexpoint, manip_as_in_sim, hermes}. Moreover, grasp synthesis models typically rely on 3D representation to model the scene~\citep{anygrasp, dexgraspnet2, drograsp, dexgraspnet, gendexgrasp, unidexgrasp, ultradexgrasp}, thereby enabling 
the prediction of grasp poses grounded in 3D geometry. Collision-free motion planning in real-world environments also requires 3D representation~\citep{curobo, neuralmp, drp, pyroki}, as explicit 3D data provide superior spatial and geometric information for obstacle avoidance. 
Recently, several studies leverage implicit 3D representation to enhance the performance of manipulation policies~\citep{spatial_forcing, gp3, gvftape, evo0}.
We evaluate the quality and usability of the 3D representation reconstructed by \ours in downstream applications such as imitation learning, sim-to-real transfer, grasp synthesis, and collision-free motion planning.

\section{Methodology}
\label{sec:method}

An overview of \ours is shown in Fig.~\ref{fig:model}. In this section, we describe \ours in detail. We begin with a brief formulation of the task and 3D representation (Section \ref{sec:method_formulation}). Next, we elaborate on the carefully designed model architecture (Section \ref{sec:method_model}). We then discuss the curation of Robo3R-4M, a large-scale, high-quality synthetic dataset that supports the training of \ours (Section \ref{sec:method_dataset}). Finally, we provide details about the training objectives (Section \ref{sec:method_training_objectives}).

\subsection{Formulation}
\label{sec:method_formulation}

\subsubsection{Task Definition}
The task we address is metric-scale and fine-grained 3D reconstruction from sparse views in robotic manipulation scenarios. Specifically, the input consists of monocular or binocular RGB images $\{I_i\}_{i=1}^N, I_i \in \mathbb{R}^{H \times W \times 3}, N \in \{1, 2\}$, and the robot state $\mathbf{J} \in \mathbb{R}^Q$, where $\mathbf{J}$ denotes the robot’s joint angles and $Q$ is the number of joints. A feed-forward neural network performs the 3D reconstruction and outputs a comprehensive set of 3D attributes, including depth maps $D \in \mathbb{R}^{H \times W}$, normalized image coordinates $C \in \mathbb{R}^{H \times W \times 2}$ (which represent the 2D pixel positions projected onto the normalized image plane at unit depth), relative camera translation $\mathbf{t}_{\text{rel}} \in \mathbb{R}^{3}$, relative camera rotation $\mathbf{R}_{\text{rel}} \in \mathbb{R}^{3 \times 3}$, and a global similarity transformation $\mathbf{S} \in \mathbb{R}^{4 \times 4}$.

\subsubsection{Scale-Invariant Local 3D Representation}

Directly predicting metric-scale 3D points in the world coordinate system is challenging. Therefore, we first predict a scale-invariant local 3D representation in the camera coordinate system. Given normalized image coordinates $\mathbf{c} = (x, y) \in \mathbb{R}^2$ and depth $d \in \mathbb{R}$, we derive the local point cloud $\mathbf{P}_{\text{local}} \in \mathbb{R}^3$ via unprojection. Since the normalized coordinates represent the 2D projection on the unit depth plane ($d=1$), the 3D position can be recovered by scaling the direction vector $(x, y, 1)$ by the depth $d$:
\begin{equation}
\mathbf{P}_{\text{local}} = [ x \cdot d, \ y \cdot d, \ d ]^\top.
\end{equation}
This formulation decouples ray direction from depth estimation, enabling the network to learn geometric structure in a local canonical space before transforming it to the global frame.
To address scale ambiguity, \ours predicts scale-invariant local points. Specifically, before computing the loss between the predicted local points $\hat{\mathbf{P}}_{\text{local}}$ and the ground truth $\mathbf{P}_{\text{local}}^{\text{gt}}$, we multiply $\hat{\mathbf{P}}_{\text{local}}$ by a scale factor $s$ that is determined by aligning the prediction with the ground truth.

\subsubsection{Metric-Scale Geometry in the Canonical Frame}

After obtaining the scale-invariant local points $\mathbf{P}_{\text{local}}$, we register the local points from multiple views $\{\mathbf{P}^i_{\text{local}}\}_{i=1}^N$ via the predicted relative camera translation $\mathbf{t}_{\text{rel}}$ and rotation $\mathbf{R}_{\text{rel}}$:

\begin{equation}
\mathbf{P}_{\text{reg}} = \left\{\, \mathbf{R}_{\text{rel}}^i \mathbf{P}^i_{\text{local}} + \mathbf{t}_{\text{rel}}^i \mid i = 1, \ldots, N \right\},
\end{equation}
where $i$ indexes the $N$ different views.
Then, we transform the registered points $\mathbf{P}_{\text{reg}}$ into the metric-scale geometry in the canonical robot frame $\mathbf{P}_{\text{cano}}$ via the predicted global similarity transformation $S$:
\begin{equation}
\mathbf{P}_{\text{cano}} = \left\{\, \left[\, \mathbf{p};\, 1\, \right] \mathbf{S}^{\top} \mid \mathbf{p} \in \mathbf{P}_{\text{reg}} \right\}.
\end{equation}

\subsection{Model Architecture}
\label{sec:method_model}

\begin{figure}[t!]
    \centering
    \includegraphics[width=0.98\linewidth]{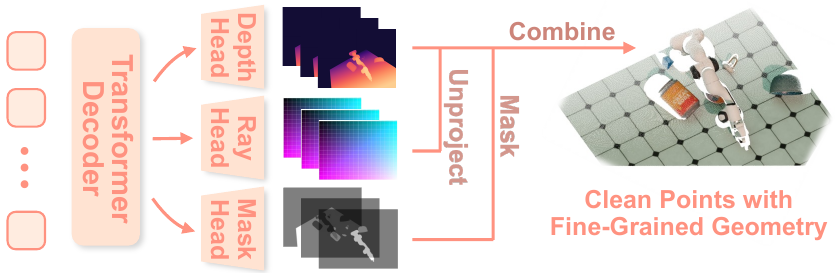}
    \caption{\textbf{Masked point head.}
    To address the over-smoothing problem for dense prediction, we propose a masked point head that decomposes point prediction into depth, normalized image coordinate, and mask predictions. Through unprojection, masking, and combination, we obtain sharp points with fine-grained geometric details.}
    \label{fig:masked_point_head}
    \vspace{-1mm}
\end{figure}

\begin{figure}[t!]
    \centering
    \includegraphics[width=0.98\linewidth]{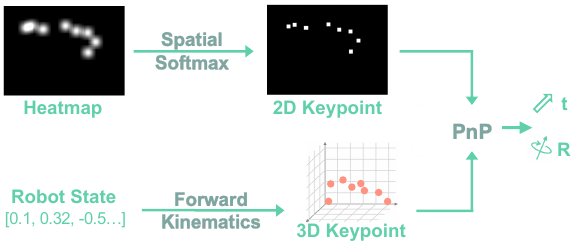}
    \caption{\textbf{Extrinsic estimation module.}
    The extrinsic estimation module extracts robot keypoints and accurately estimates the camera extrinsics by solving the Perspective-n-Point (PnP) problem; the camera extrinsics are used to refine the global similarity transformation.}
    \label{fig:extrinsic_estimation}
    \vspace{-3mm}
\end{figure}


\subsubsection{Encoders}
The inputs to \ours are images and robot states.
We employ DINOv2 ViT-L to encode images from $N$ views $\{I_i\}_{i=1}^N, I_i \in \mathbb{R}^{H \times W \times 3}$ into patch features $\{F_{I,i}\}_{i=1}^N, F_{I,i} \in \mathbb{R}^{\frac{H}{14} \times \frac{W}{14} \times 1024}$. The robot states are projected to state features $F_J \in \mathbb{R}^{1024}$ using a multilayer perceptron (MLP) with GeLU activations. The image and state features are then fused via element-wise addition to obtain the combined features $F \in \mathbb{R}^{N \times \frac{H}{14} \times \frac{W}{14} \times 1024}$. Then, we append learnable similarity transformation (S.T.) tokens to $F$, which serve as the input to the transformer backbone.

\subsubsection{Transformer Backbone}
We employ a transformer backbone based on the Alternating-Attention mechanism~\citep{vggt}. Specifically, we stack 18 alternating global and frame-wise attention blocks to enable efficient propagation of information both within and across frames.

\subsubsection{Prediction Heads}
~

\noindent\textbf{Masked point head.}
Over-smoothing is a persistent issue in dense prediction, which could result in blurry edges and loss of detail in point clouds. 
To address this problem, we decouple dense point prediction into depth, image coordinate, and mask components, as illustrated in Fig.~\ref{fig:masked_point_head}. 
The masked point head consists of three branches dedicated to the robot, objects, and background. Each branch contains a depth head, a ray head, and a mask head, and produces a point map by unprojecting the depth along the corresponding ray directions. The robot, object, and background masks then extract region-specific points from their respective point maps, which are aggregated to reconstruct the full scene point cloud.
Specifically, the masked point head consists of a five-layer transformer decoder, a depth head, a ray head, and a mask head. Patch-wise features produced by the transformer backbone are processed by the transformer decoder.
The outputs are then decoded into depth, image coordinates, and masks using the MLP heads followed by pixel shuffle operations.

\noindent\textbf{Relative pose head.}
Adapted from~\citet{reloc3r} and ~\citet{pi3}, the head processes outputs of the transformer backbone with a transformer decoder, two residual convolution blocks and adaptive average pooling, followed by an MLP. It predicts translation as a 3D vector and rotation in a 9D representation. The 9D rotation is reshaped into a $3 \times 3$ matrix and orthogonalized via SVD, ensuring a valid rotation.

\noindent\textbf{Similarity transformation head.}
The latent representation of the similarity transformation extracted by the S.T. tokens is decoded by a network with the same architecture as the relative pose head, together with an MLP that predicts the scale. This process yields the rigid transformation $\mathbf{T}$ and the scale $s$, from which the similarity transformation $\mathbf{S}$ can be obtained:
\begin{equation}
    \mathbf{S} = s \cdot \mathbf{T}.
\end{equation}

\noindent\textbf{Extrinsic estimation module and keypoint head.}
Although \ours directly predicts the camera extrinsics, which are part of the similarity transformation, we additionally develop a more accurate and robust extrinsic estimation module, as illustrated in Fig.~\ref{fig:extrinsic_estimation}. The results obtained from this module can further refine the similarity transformation. 
We compute 3D keypoints as the frame origin of each link by applying forward kinematics to the robot's joint positions. During data synthesis, these keypoints are projected onto the image plane via camera intrinsics and extrinsics to obtain 2D keypoints, each rendered as a Gaussian blob to produce a per-keypoint heatmap. 
During inference, Soft-Argmax is applied to heatmaps to extract 2D keypoints as a weighted average of pixel coordinates, yielding floating-point values that enable sub-pixel accuracy.
The 3D keypoints and per-keypoint heatmaps are index-aligned, enabling matching between 3D and 2D keypoints. 
\ours predicts the pixel coordinates of these keypoints with a keypoint head, and the camera extrinsics are subsequently estimated by solving a PnP problem. In the keypoint head, patch features are first processed by a transformer decoder, then decoded by an MLP followed by a pixel shuffle operation to generate a keypoint heatmap $M_{kp} \in \mathbb{R}^{H \times W \times N_{kp}}$, where $N_{kp}$ is the number of keypoints. Finally, we employ a differentiable Soft-Argmax operator to extract 2D keypoint coordinates $C_{kp} \in \mathbb{R}^{N_{kp} \times 2}$ from the heatmaps with sub-pixel accuracy.

\subsection{Synthetic Data Pipeline}
\label{sec:method_dataset}

\begin{figure}[t!]
    \centering
    \includegraphics[width=\linewidth]{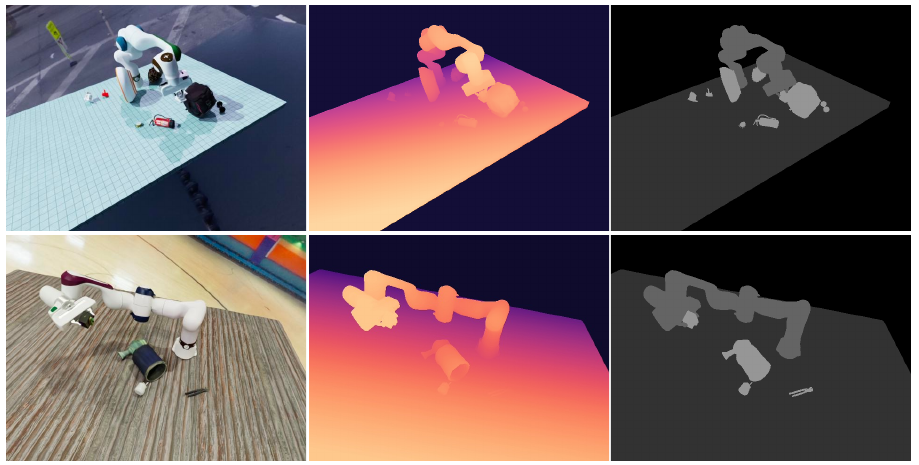}
    \caption{\textbf{Data samples.} The dataset showcases a diverse array of assets with extensive randomization, encompassing rich modalities and comprehensive annotations.}
    \label{fig:data_sample}
    \vspace{-3mm}
\end{figure}

We curate a large-scale synthetic dataset, Robo3R-4M, using NVIDIA Isaac Sim as both the physics and rendering engine. 
We utilize a diverse set of assets, including 16,911 objects sourced from DTC~\citep{dtc} and Objaverse~\citep{objaverse}, 4,710 textures, and 6,512 environment maps.
Extensive domain randomization is applied to various factors, including robot behaviors, camera extrinsics and intrinsics, object instances and poses, table instances and poses, background, and lighting conditions. Multiple data modalities are recorded, including RGB images, depth images, semantic masks, robot states, camera intrinsics and extrinsics. Representative data samples are illustrated in Fig.~\ref{fig:data_sample}. Robo3R-4M consists of 100,000 scenes and four million photorealistic, high-quality frames, which support the training of \ours. See more details about synthetic data generation in Appendix~\ref{appendix:synthetic_data_generation_details}.

\subsection{Training Objectives}
\label{sec:method_training_objectives}

\noindent\textbf{Point loss.}
Instead of directly supervising the depth and normalized image coordinates, we supervise the point cloud obtained via unprojection. The point loss is defined as follows:
\begin{equation}
\mathcal{L}_{\text{point}} = \frac{1}{3HW} \sum_{i=1}^{H \times W} \left\| s \cdot \hat{\mathbf{p}}_{i} - \mathbf{p}_{i} \right\|_1,
\end{equation}
where $s$ is a scale factor determined by aligning the predicted point maps with the ground truth.

\noindent\textbf{Normal loss.}
To further supervise the geometric consistency of the predicted point clouds, we supervise surface normals during training. 
For each point in the predicted point map, the normal vector is computed by taking the cross product of vectors connecting the point to its neighboring points. The normal loss is defined as:
\begin{equation}
\mathcal{L}_{\text{normal}} = \frac{1}{K} \sum_{k=1}^{K} \Delta \theta_k,
\end{equation}
where $K$ denotes the number of valid normal pairs, and $\Delta \theta_k$ is the angle between the predicted and ground-truth normals at location $k$. The angle is computed as
\begin{equation}
\Delta \theta_k = \arctan2 \left( \left\| \hat{\mathbf{n}}_k \times \mathbf{n}_k \right\|,\, \hat{\mathbf{n}}_k \cdot \mathbf{n}_k \right),
\end{equation}
where $\hat{\mathbf{n}}_k$ and $\mathbf{n}_k$ denote the normal vectors estimated from the predicted and ground-truth points, respectively.

\noindent\textbf{Mask loss.}
In the masked point head, we employ pixel-wise masks to obtain clean points with fine-grained geometry. Specifically, we predict separate masks for the robot, objects, and background. The mask loss is defined as the binary cross-entropy between the predicted masks $\hat{m}_i$ and the ground-truth masks $m_i$:
\begin{equation}
\mathcal{L}_{\text{mask}} = \frac{1}{HW} \sum_{i=1}^{H \times W} \mathrm{BCE}\left(\hat{m}_i,\, m_i\right).
\end{equation}

\noindent\textbf{Relative pose loss.}
The relative pose head predicts a camera translation $\hat{\mathbf{t}}$ and rotation $\hat{\mathbf{R}}$ for each view. The relative pose from view $j$ to view $i$ is computed as:
\begin{equation}
\hat{\mathbf{R}}_{\text{rel}} = \hat{\mathbf{R}}_i^{-1} \hat{\mathbf{R}}_j,
\end{equation}
\begin{equation}
\hat{\mathbf{t}}_{\text{rel}} = \hat{\mathbf{R}}_i^{-1} (\hat{\mathbf{t}}_j - \hat{\mathbf{t}}_i),
\end{equation}
where $\hat{\mathbf{R}}_i$ and $\hat{\mathbf{t}}_i$ denote the predicted rotation and translation of view $i$. The relative pose loss is defined as:
\begin{equation}
\mathcal{L}_{\text{rel}} = \alpha \cdot \mathrm{Huber}\left(\hat{\mathbf{t}}_{\text{rel}},\, \mathbf{t}_{\text{rel}} \right) + \mathrm{Angle}\left(\hat{\mathbf{R}}_{\text{rel}},\, \mathbf{R}_{\text{rel}}\right),
\end{equation}
where $\mathrm{Huber}(\cdot, \cdot)$ denotes the Huber loss for translation, $\mathrm{Angle}(\cdot, \cdot)$ computes the rotation angular error, and $\alpha$ is a balancing weight.

\noindent\textbf{Similarity transformation loss.}
The similarity transformation head predicts the global similarity transformation, which consists of a scale factor $\hat{s}$, as well as the camera translation $\hat{\mathbf{t}}_{\text{abs}}$ and rotation $\hat{\mathbf{R}}_{\text{abs}}$ for the first view. The similarity transformation loss supervises these components and is defined as:
\begin{align}
\mathcal{L}_{\text{ST}} =\ 
&\beta_1 \cdot \mathrm{Huber}\left(\hat{s},\, s \right) \notag \
+ \beta_2 \cdot \mathrm{Huber}\left(\hat{\mathbf{t}}_{\text{abs}},\, \mathbf{t}_{\text{abs}} \right) \notag \\
&+ \mathrm{Angle}\left(\hat{\mathbf{R}}_{\text{abs}},\, \mathbf{R}_{\text{abs}}\right).
\end{align}

\noindent\textbf{Keypoint loss.}
The keypoint loss jointly supervises the keypoint heatmap and the 2D keypoint coordinates, which are extracted from the heatmap using a differentiable Soft-Argmax operator. The loss is defined as:
\begin{equation}
\mathcal{L}_{\text{kp}} = \gamma \cdot \left\| \hat{M}_{\text{kp}} - M_{\text{kp}} \right\|_1 + \left\| \hat{C}_{\text{kp}} - C_{\text{kp}} \right\|_1,
\end{equation}
where $\hat{M}_{\text{kp}}$ and $M_{\text{kp}}$ denote the predicted and ground-truth keypoint heatmaps, $\hat{C}_{\text{kp}}$ and $C_{\text{kp}}$ are the predicted and ground-truth keypoint coordinates, and $\gamma$ is balancing weights.

\noindent\textbf{Overall.}
\ours is trained end-to-end by minimizing the multi-task loss $\mathcal{L}$, which is the weighted sum of the point loss, normal loss, mask loss, relative pose loss, similarity transformation loss, and keypoint loss:
\begin{align}
\mathcal{L} =\ 
&\lambda_1 \mathcal{L}_{\text{point}} 
+ \lambda_2 \mathcal{L}_{\text{normal}} 
+ \lambda_3 \mathcal{L}_{\text{mask}} 
\notag \\
&+ \lambda_4 \mathcal{L}_{\text{rel}} 
+ \lambda_5 \mathcal{L}_{\text{ST}} 
+ \lambda_6 \mathcal{L}_{\text{kp}},
\end{align}
where $\lambda_1, \ldots, \lambda_6$ are the weights for each loss term. Additional training details are provided in Appendix~\ref{appendix:training_details}.

\section{Experiments}
\label{sec:experiments}

We conduct extensive experiments to evaluate the effectiveness of \ours, with respect to both reconstruction quality and performance on downstream tasks. Specifically, we seek to address the following questions:

\begin{enumerate}
\renewcommand{\labelenumi}{\arabic{enumi})}

\item How does \ours perform in terms of 3D reconstruction quality? In particular, does it achieve high-fidelity geometry, and does it accurately predict scale and camera pose? 

\item How robust is \ours in challenging scenarios where depth cameras typically fail?

\item Can \ours be successfully applied to various downstream robotic manipulation applications? 

\item Do the key design components effectively improve the performance of reconstruction?
\end{enumerate}

\subsection{3D Reconstruction Quality}

\subsubsection{Quantitative Performance}
~

\noindent\textbf{Benchmark.}
We constructed a benchmark for 3D reconstruction in robotic manipulation scenarios by rendering a photorealistic and diverse dataset, using objects, textures, and environment maps that are different from those used in training.
The test set contains 2,000 scenes and 80,000 frames. 
The evaluation consists of two aspects: point map estimation and relative camera pose estimation. The metrics for point map estimation include scale-invariant point error and scale error. For relative camera pose estimation, the metrics include relative translation error (RTE), relative rotation error (RRE), relative translation accuracy (RTA) at a given threshold (e.g., RTA@0.03 for 0.03 meters), and relative rotation accuracy (RRA). As robots typically perceive the environment from sparse views, we evaluate point map estimation under monocular and binocular settings, and assess relative camera pose estimation under the binocular setting.

\noindent\textbf{Baselines.}
We select four leading feed-forward 3D reconstruction models as baselines: VGGT~\citep{vggt}, $\pi^3$~\citep{pi3}, DepthAnything3 (DA3)~\citep{da3}, and MapAnything (MA)~\citep{mapanything}. Among them, MA and DA3 perform metric-scale reconstruction. Additionally, we consider MapAnything fine-tuned on the Robo3R-4M dataset (MA-FT) as a baseline.

\noindent\textbf{Results.}
We evaluate \ours against strong baselines on the benchmark. 
Tab.~\ref{table:exp_recon_point_map} presents the quantitative results for point map estimation. Our method significantly outperforms all baselines across all metrics in both monocular and binocular settings. In the monocular case, \ours achieves a point error of 0.006, representing an order of magnitude improvement over the second-best method, $\pi^3$. Unlike the other models, which suffer from severe scale ambiguity (scale errors $>0.46$), \ours effectively recovers metric geometry. This superiority persists in the binocular setting, where \ours further reduces errors, demonstrating robustness in 3D reconstruction for challenging manipulation scenarios.
Additionally, \ours consistently outperforms the fine-tuned MapAnything model (MA-FT). 
This highlighting the benefits of our proposed model architecture.
As shown in Tab.~\ref{table:exp_recon_camera_pose}, our method demonstrates exceptional precision in relative pose estimation. \ours achieves an RTE of 0.014 and an RRE of 0.013, which are approximately $8\times$ and $5\times$ lower than those of the best baseline, $\pi^3$, respectively. Furthermore, the high accuracy rates (RTA@0.03 of 0.951) confirm that \ours consistently provides reliable relative camera pose.

\begin{table}[t!]
    \centering
    \caption{\textbf{Results on point map estimation.} We report the scale-invariant point error and scale error for each method. 
    }
    \label{table:exp_recon_point_map}

    \begin{tabular}{l|l|cc}
    \toprule
    & Method & Point Err. $\downarrow$ & Scale Err. $\downarrow$ \\
    \midrule

    \multirow{4}{*}{Monocular} 
    & VGGT & 0.126 & 0.663 \\
    & $\pi^3$ & 0.061 & 0.497 \\
    & DA3 & 0.075 & 0.506 \\
    & MA & 0.078 & 0.467 \\
    & MA-FT & 0.010 & 0.010 \\
    & Ours & \cellcolor{mygreen_light}\textbf{0.006} & \cellcolor{mygreen_light}\textbf{0.007} \\

    \midrule

    \multirow{4}{*}{Binocular} 
    & VGGT & 0.220 & 0.619 \\
    & $\pi^3$ & 0.032 & 0.483 \\
    & DA3 & 0.042 & 0.719 \\
    & MA & 0.076 & 0.540 \\
    & MA-FT & 0.009 & 0.007 \\
    & Ours & \cellcolor{mygreen_light}\textbf{0.005} & \cellcolor{mygreen_light}\textbf{0.004} \\

    \bottomrule
    \end{tabular}
    \vspace{-1mm}

\end{table}

\begin{table}[t!]
    \centering
    \caption{\textbf{Results on relative camera pose prediction.} We report the relative translation error (RTE), relative rotation error (RRE), relative translation accuracy (RTA), and relative rotation accuracy (RRA) for each method. 
    }
    \label{table:exp_recon_camera_pose}
    \begin{tabular}{l|cccc}
    \toprule
    Method & RTE $\downarrow$ & RRE $\downarrow$ & RTA@0.03 $\uparrow$ & RRA@0.03 $\uparrow$ \\
    \midrule
    VGGT & 0.373 & 0.316 & 0.014 & 0.052 \\
    $\pi^3$ & 0.116 & 0.073 & 0.110 & 0.245 \\
    MA & 0.168 & 0.116 & 0.031 & 0.069 \\
    DA3 & 0.160 & 0.111 & 0.129 & 0.226 \\
    Ours & \cellcolor{mygreen_light}\textbf{0.014} & \cellcolor{mygreen_light}\textbf{0.013} & \cellcolor{mygreen_light}\textbf{0.951} & \cellcolor{mygreen_light}\textbf{0.899} \\
    \bottomrule
    \end{tabular}
    \vspace{-3mm}

\end{table}

\subsubsection{Qualitative Comparisons in Real-World Scenarios}

\begin{figure*}[ht]
    \centering
    \includegraphics[width=\linewidth]{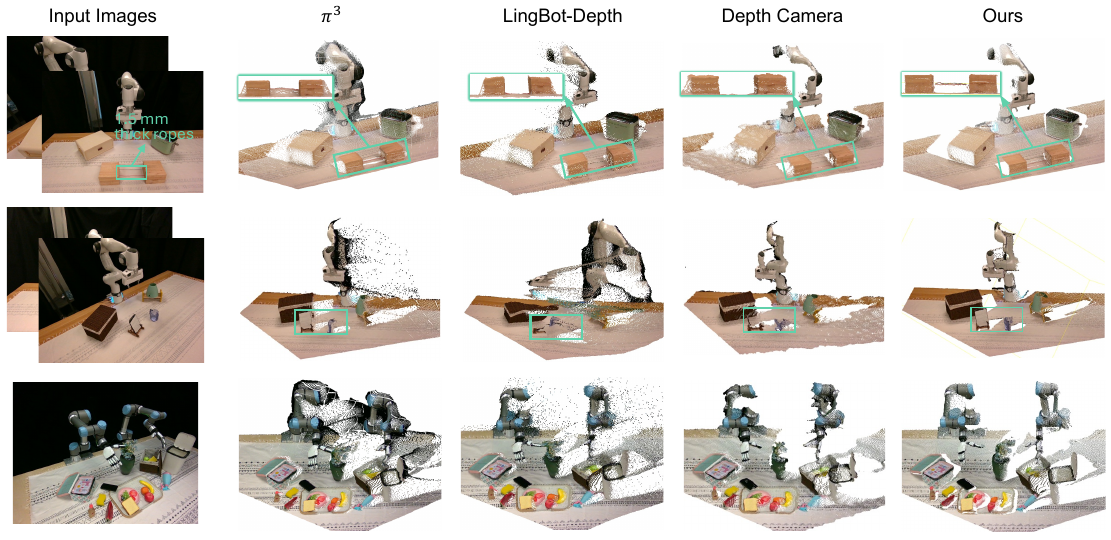}
    \caption{\textbf{Qualitative Comparisons of 3D geometry.} We use a RealSense D455 as the depth camera and manually align the scale for point clouds reconstructed by $\pi^3$. We evaluate the methods on several challenging scenarios, including a tiny object (row 1), a scene with a mirror and a transparent cup (row 2), and a cluttered environment (row 3).}
    \label{fig:qualitative_comparison}
    \vspace{-3mm}
\end{figure*}

We compare our method with the state-of-the-art feed-forward 3D reconstruction model $\pi^3$~\citep{pi3}, the depth completion model LingBot-Depth~\citep{lingbot-depth}, as well as the depth camera.
Compared to the other reconstruction models and the depth camera, \ours achieves significantly cleaner, more accurate 3D geometry with finer-grained details, and robustly handles challenging scenarios where the other methods fail, 
as illustrated in Fig.~\ref{fig:qualitative_comparison}. Specifically, \ours is capable of reconstructing objects as narrow as 1.5~mm (spanning only 1 to 2 pixels in the image), whereas other methods, including depth cameras, fail to capture such fine geometry (row 1). Furthermore, \ours successfully handles reflective and transparent objects that blind depth sensors (row 2). Even in cluttered scenes that include bimanual robots with dexterous hands, \ours consistently produces accurate and clean point clouds (row 3).

\subsection{Downstream Robotic Manipulation}

\noindent\textbf{Overview of the manipulation benchmark.}
We evaluate the effectiveness of \ours for robotic manipulation on the real-world benchmark. 
The robotic platforms include a single-arm Franka Research 3 with a parallel gripper and a bimanual UR5e, each equipped with an XHand. RealSense D455 cameras are used to acquire RGB and depth images of the scene. \ours and the manipulation policy are deployed on an NVIDIA RTX 4090. The detailed inference speed of \ours is provided in Appendix~\ref{appendix:model_size_and_inference_speed}. The control frequency of the robotic system is set to 10 Hz. We conduct experiments across four applications: imitation learning, sim-to-real transfer, grasp synthesis, and collision-free motion planning. Additional details regarding the real-world manipulation experiments are provided in Appendix~\ref{appendix:real-world_experiment_details}.

\subsubsection{Imitation Learning}
\label{section:exp_imitation_learning}
~

\noindent\textbf{Experimental setup and baselines.}
\ours is evaluated on four tasks:

\begin{enumerate}
    \item \textbf{Sweep Bean}: The gripper grasps the handle of a broom and sweeps beans on the table into a dustpan. The small size of the beans presents challenges for depth sensors and scene reconstruction.
    \item \textbf{Insert Screw}: The robot inserts a screw into a hole where the hole’s radius exceeds the screw’s radius by only 2~mm, requiring millimeter-level reconstruction and manipulation precision.
    \item \textbf{Breakfast}: This is a long-horizon task in which the robot first picks up a slice of bread, places it in a toaster, presses the toaster button to toast the bread, then picks up a milk jug, pours milk into a glass, returns the jug to its original position, and finally pushes the glass forward.
    \item \textbf{BiDex Pour}: This is a bimanual dexterous manipulation task, where the left hand holds a cup and the right hand holds a kettle to pour water into the cup.
\end{enumerate}
We use Maniflow (MF)~\citep{maniflow} as both a 2D- and 3D-based manipulation policy. \ours takes RGB images as input and produces point clouds, which are then used as input for the 3D-based MF. The baseline is the 2D-based MF, which takes RGB images as input. Additionally, as baselines, we use point clouds reconstructed by the other feed-forward 3D reconstruction models (Other FFs), as well as those obtained from a depth camera, as inputs for the 3D-based MF. $\pi_0$~\citep{pi0}, a state-of-the-art vision-language-action model, is also selected as a baseline. 

\begin{figure}[t!]
    \centering
    \includegraphics[width=\linewidth]{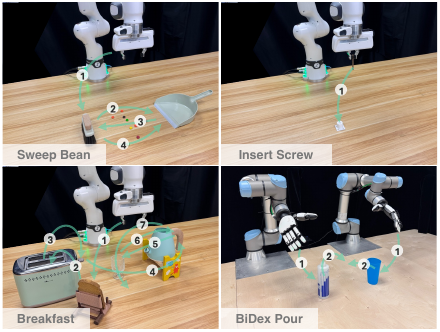}
    \caption{\textbf{Imitation learning tasks.} We design four manipulation tasks for imitation learning evaluation: Sweep Bean, Insert Screw, Breakfast, and BiDex Pour. 
    }
    \label{fig:task_illustration}
    \vspace{-3mm}
\end{figure}

\noindent\textbf{Results.}
As can be seen in Tab.~\ref{table:exp_mani_real}, \ours combined with the 3D-based MF demonstrates superior or competitive performance compared to all baselines.
Across all tasks, \ours (MF + Ours) consistently outperforms the 2D-based MF baseline and $\pi_0$, confirming the benefit of lifting 2D observations into 3D for manipulation policies. Furthermore, \ours is better than depth cameras in scenarios involving small objects or high precision. The ``Other FFs'' baseline failed to produce feasible actions (indicated by ``-''), due to significant errors in scale and accuracy of the geometry, as well as the inability to effectively crop out cluttered background.



\begin{table}[t!]
    \centering
    \caption{\textbf{Results on imitation learning.} We report the number of successes over the total number of trials. 
    }
    \label{table:exp_mani_real}
    \begin{tabular}{l|cc}
        \toprule
        Method & Sweep Bean & Insert Screw \\
        \midrule
        $\pi_0$ & 11/16 & 4/16 \\
        MF + RGB Camera & 10/16 & 2/16 \\
        MF + Other FFs & - & - \\
        MF + Depth Camera & 4/16 & 7/16 \\
        MF + Ours & \cellcolor{mygreen_light}\textbf{14/16} & \cellcolor{mygreen_light}\textbf{15/16} \\
        \midrule\midrule
        Method & Breakfast & BiDex Pour \\
        \midrule
        $\pi_0$ & 4/16 & 12/16 \\
        MF + RGB Camera & 5/16 & 9/16 \\
        MF + Other FFs & - & - \\
        MF + Depth Camera & 11/16 & \cellcolor{mygreen_light}\textbf{16/16} \\
        MF + Ours & \cellcolor{mygreen_light}\textbf{12/16} & \cellcolor{mygreen_light}\textbf{16/16} \\
        \bottomrule
    \end{tabular}
    \vspace{-2mm}
    
\end{table}

\begin{table}[t!]
    \centering
    \caption{\textbf{Results on sim-to-real transfer.} The number of successes over the total number of trials is reported. 
    }
    \label{table:exp_mani_sim2real}
    \begin{tabular}{l|cc}
    \toprule
    Method & Push Cube & Pick Cube \\
    \midrule
    RGB Camera & 3/16 & 2/16 \\
    Depth Camera & 7/16 & 5/16 \\
    Ours & \cellcolor{mygreen_light}\textbf{16/16} & \cellcolor{mygreen_light}\textbf{12/16} \\
    \bottomrule
    \end{tabular}
    \vspace{-3mm}

\end{table}

\begin{table}[ht]
    \centering
    \caption{\textbf{Results on grasp synthesis.} The number of successes over the total number of trials is reported. 
    }
    \label{table:exp_mani_grasp_synthesis}
    \begin{tabular}{l|ccc}
    \toprule
    Method & Normal & Transparent or Reflective & Small \\
    \midrule
    Other FFs & - & - & - \\
    Depth Camera & 12/16 & 7/16 & 6/16 \\
    Ours & \cellcolor{mygreen_light}\textbf{14/16} & \cellcolor{mygreen_light}\textbf{10/16} & \cellcolor{mygreen_light}\textbf{11/16} \\
    \bottomrule
    \end{tabular}
    \vspace{-1mm}

\end{table}
\begin{table}[t!]
    \centering
    \caption{\textbf{Results on collision-free motion planning.} The number of successes over the total number of trials is reported. 
    }
    \label{table:exp_mani_collision_free_mp}
    \begin{tabular}{l|ccc}
    \toprule
    Method & Normal & Transparent or Reflective & Thin \\
    \midrule
    Other FFs & - & - & - \\
    Depth Camera & \cellcolor{mygreen_light}\textbf{5/5} & 2/5 & 2/5 \\
    Ours & \cellcolor{mygreen_light}\textbf{5/5} & \cellcolor{mygreen_light}\textbf{4/5} & \cellcolor{mygreen_light}\textbf{5/5} \\
    \bottomrule
    \end{tabular}
    \vspace{-3mm}

\end{table}

\subsubsection{Sim-to-Real Transfer}
~

\noindent\textbf{Experimental setup and baselines.}
We collect 200 demonstrations for each of the ``Push Cube'' and ``Pick Cube'' tasks in NVIDIA Isaac Sim.
The baseline methods acquire data using either RGB or depth cameras in simulation and deploy policies using the same modality in the real world. \ours reconstructs 3D geometry from RGB images in both simulation and the real world, using the resulting point clouds as inputs for policy training and deployment.

\noindent\textbf{Results.}
The visualization of the visual gap between simulation and reality for different methods is presented in Appendix~\ref{appendix:real-world_experiment_details}. RGB images contain redundant information such as high-frequency textures, and the depth in simulation is perfect, whereas real-world depth data are inaccurate and noisy. As a result, directly using RGB, depth, or point cloud data generally leads to a significant sim-to-real gap. \ours achieves a consistent scene representation across domains and significantly reduces the sim-to-real gap. In both tasks, \ours achieves a notably higher success rate compared to the baselines, as shown in Tab.~\ref{table:exp_mani_sim2real}.

\subsubsection{Grasp Synthesis}
\label{subsection:grasp_synthesis}
~

\noindent\textbf{Experimental setup and baselines.}
We employ AnyGrasp~\citep{anygrasp} as the grasp synthesis model, using point clouds reconstructed by \ours as its input. As baselines, we use point clouds obtained from a depth camera and from the other feed-forward 3D reconstruction models as inputs to AnyGrasp. We conduct experiments on normal, transparent, reflective, and small objects.

\noindent\textbf{Results.}
The results are presented in Tab.~\ref{table:exp_mani_grasp_synthesis}. The other feed-forward 3D reconstruction models fail to enable AnyGrasp to produce feasible grasps (indicated by ``-''). 
While the depth camera baseline achieved reasonable performance on normal objects (12/16), it struggled with challenging scenarios, dropping to 7/16 for transparent/reflective objects and 6/16 for small objects. In contrast, our method demonstrated superior performance, achieving the highest success rates in all categories.
These results confirm that \ours significantly improves grasp synthesis robustness, particularly for objects with challenging materials or fine-grained geometries.

\subsubsection{Collision-Free Motion Planning}
~

\noindent\textbf{Experimental setup and baselines.}
We use cuRobo~\citep{curobo} as motion planner.
The selection of baselines follows Sec.~\ref{subsection:grasp_synthesis}.

\noindent\textbf{Results.}
Consistent with the grasp synthesis experiments, the other feed-forward 3D reconstruction models failed to enable successful obstacle avoidance. While the depth camera proves effective for normal objects, it frequently fails to capture the geometry of challenging obstacles, resulting in a low success rate of 2/5 for transparent, reflective, and thin objects. 
In contrast, our method maintains high reliability across all scenarios, as shown in Tab.~\ref{table:exp_mani_collision_free_mp}. 
These results highlight that the accurate geometry provided by \ours is critical for avoiding collisions with objects that are invisible to depth sensors.

\subsection{Effectiveness of Design Choices}
We validate the effectiveness of the proposed extrinsic estimation module (KP + PnP) by comparing it against direct extrinsic prediction (Direct Pred.). As shown in Tab.~\ref{table:exp_ablation_cam}, the extrinsic estimation module yields lower absolute translation and rotation errors, as well as higher absolute translation and rotation accuracy.
These results confirm that predicting keypoints followed by a PnP solver provides a more robust and precise estimation of the camera pose compared to directly regressing the extrinsics.

We also investigate the contributions of incorporating robot state during feed-forward reconstruction. The results in Tab.~\ref{table:exp_ablation_robot_state} indicate that fusing robot states with image features via element-wise addition not only yields better point map estimation compared to methods that do not use robot states, but also achieves more accurate absolute camera pose in global similarity transformation. Furthermore, this approach outperforms other modality fusion methods, such as employing self-attention within the transformer backbone.

\begin{figure}[ht]
    \centering
    \includegraphics[width=0.98\linewidth]{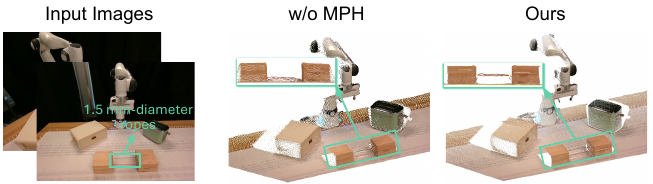}
    \caption{\textbf{Reconstruction w/ and w/o the masked point head.}}
    \label{fig:qualitative_comparison_ablation_mph}
    \vspace{-7mm}
\end{figure}

We provide qualitative comparisons for the masked point head, since its advantage in producing finer-grained geometry is not well captured by conventional benchmarks. 
Specifically, we train a variant that retains the point decoder architecture but removes the masked design (w/o MPH). As shown in Fig.~\ref{fig:qualitative_comparison_ablation_mph}, the masked point head enables the model to predict clean points on structures as fine as 1.5 mm, whereas the baseline fails to recover such geometry.

\begin{table}[t!]
    \centering
    \caption{\textbf{Ablation study of camera pose prediction in the robot base frame.} We report absolute translation error (ATE), absolute rotation error (ARE), absolute translation accuracy at a threshold of 0.01 meters (ATA@0.01), and absolute rotation accuracy (ARA@0.01).}
    \label{table:exp_ablation_cam}

    \begin{tabular}{l|cccc}
    \toprule
    Method & ATE $\downarrow$ & ARE $\downarrow$ & ATA@0.01 $\uparrow$ & ARA@0.01 $\uparrow$ \\
    \midrule
    Direct Pred. & 0.018 & 0.018 & 0.334 & 0.359 \\
    KP + PnP & \cellcolor{mygreen_light}\textbf{0.016} & \cellcolor{mygreen_light}\textbf{0.016} & \cellcolor{mygreen_light}\textbf{0.442} & \cellcolor{mygreen_light}\textbf{0.415} \\
    \bottomrule
    \end{tabular}

\end{table}

\begin{table}[t!]
    \centering
    \caption{\textbf{Ablation study on conditioning on robot state.} The point error, normal error, absolute translation error (ATE), and absolute rotation accuracy (ARA) are reported, with both ATE and ARA evaluated at a threshold of 0.03 meters.}
    \label{table:exp_ablation_robot_state}

    \begin{tabular}{l|cccc}
    \toprule
    Method & Point Err. $\downarrow$ & Normal Err. $\downarrow$ & ATA $\uparrow$ & ARA $\uparrow$ \\
    \midrule
    w/o State & 0.006 & 0.081 & \cellcolor{mygreen_light}\textbf{0.903} & 0.831 \\
    Self Attn & 0.006 & 0.086 & 0.900 & 0.821 \\
    Ours & \cellcolor{mygreen_light}\textbf{0.005} & \cellcolor{mygreen_light}\textbf{0.079} & \cellcolor{mygreen_light}\textbf{0.903} & \cellcolor{mygreen_light}\textbf{0.838} \\
    \bottomrule
    \end{tabular}
    \vspace{-3mm}

\end{table}

\section{Conclusion} 
\label{sec:conclusion}

In this work, we introduce \ours, a feed-forward 3D reconstruction model designed for robotic manipulation. Trained on Robo3R-4M, a large-scale synthetic dataset for perception and reconstruction in robotic manipulation scenarios, \ours delivers high-fidelity depth estimation, precise camera parameter prediction, accurate metric scale, and a consistent, canonical coordinate system for 3D representation. Compared to depth cameras, \ours is more cost-effective, accurate, robust, and requires no calibration. It reliably handles various object materials and challenging scenarios, significantly enhancing downstream robotic applications such as imitation learning, sim-to-real transfer, grasp synthesis, and collision-free motion planning.

\noindent\textbf{Limitations and future work.}
Currently, \ours supports pinhole cameras and a limited range of embodiment types. Future work may involve generating new data to fine-tune \ours, enabling it to adapt to other camera models, such as fisheye and panoramic cameras, as well as an expanded range of embodiments.



\bibliographystyle{plainnat}
\bibliography{references}

\clearpage
\appendix
In this Appendix, we provide details on synthetic data generation (Appendix~\ref{appendix:synthetic_data_generation_details}), the implementation and training procedures, including training configurations and hyperparameters (Appendix~\ref{appendix:training_details}), model size and inference speed (Appendix~\ref{appendix:model_size_and_inference_speed}), real-world manipulation experiment protocols (Appendix~\ref{appendix:real-world_experiment_details}), as well as the generalization capabilities of \ours in diverse scenarios (Appendix~\ref{appendix:reconstruction_results_in_diverse_scenarios}).

\subsection{Synthetic Data Generation Details}
\label{appendix:synthetic_data_generation_details}

We leverage NVIDIA Isaac Sim as the simulation platform to generate large-scale synthetic datasets, utilizing Path Tracing to achieve high-fidelity, photorealistic rendering. The physics simulation operates at a frequency of 30 Hz ($dt \approx 0.033$s). To bridge the simulation-to-reality gap and improve the generalization ability of the model, we employ extensive domain randomization across visual sensors, lighting conditions, and physical scene properties.

\subsubsection{Camera Configuration}
The simulation is equipped with a multi-camera system yielding RGB images. We model the sensors as pinhole cameras with randomized intrinsic parameters. For each episode, we perturb the focal lengths ($f_x, f_y$) and principal points ($c_x, c_y$) of the camera intrinsic matrix $\mathbf{K}$. Additionally, we randomize the focus distance and f-number to simulate varying depth-of-field effects.

The camera extrinsics, $\mathbf{T}_{wc} \in SE(3)$, are randomized using a ``look-at'' constraint to ensure the workspace remains in view while varying the viewpoint. The camera positions are sampled from a spherical shell defined by randomized radius, azimuth, and elevation ranges.
The camera orientation is computed to look at a randomized target point within the robot's workspace, with an additional random roll rotation applied.

\subsubsection{Lighting Configuration}
The lighting system consists of three types of light sources: Dome, Sphere, and Distant lights. This setup allows us to effectively simulate a wide range of lighting conditions, as described below:
\begin{itemize}
    \item \textbf{Dome Light (Environment Map):} We utilize High Dynamic Range Images (HDRI) loaded from a predefined asset list. For each episode, a random HDRI is selected, and its intensity is sampled from $[300, 2000]$. The environment map is arbitrarily rotated around the Z-axis ($\pm 180^\circ$) and slightly tilted along X/Y axes ($\pm 20^\circ$).
    \item \textbf{Sphere Lights:} We instantiate $N_{sphere} \sim \{1, 2, 3\}$ point sources modeled as spheres with radii $r \in [0.05, 0.5]$m. These lights are placed randomly within a volume of $12 \times 12 \times 7$ meters around the scene. Their intensity is high-variance, sampled from $[2000, 8000]$, with randomized color temperatures and RGB tints.
    \item \textbf{Distant Lights:} We simulate directional sunlight using $N_{dist} \sim \{1, 2, 3\}$ distant light sources. These are randomized in color, intensity, and orientation to achieve diverse illumination and cast varied shadows.
\end{itemize}

\subsubsection{Scene Composition and Material Randomization}
The scene consists of a robot manipulator, manipulable objects, and background elements such as a tabletop.
\begin{itemize}
    \item \textbf{Robot:} The robot's joint configuration is initialized via an inverse kinematics (IK) solver to reach random valid end-effector poses. We apply material randomization to the robot's visual mesh, adding jitter to the metallic, roughness, and diffuse color parameters to prevent overfitting to specific robot textures.
    \item \textbf{Manipulable Objects:} Objects are loaded from Universal Scene Description (USD) assets with randomized scales. Their visual properties are varied between standard Physically Based Rendering (PBR) materials (randomized albedo, roughness, metallic) and glass materials (randomized Index of Refraction and frosting), so that the dataset includes a diverse range of materials, including transparent and reflective ones.
    \item \textbf{Background Elements:} Background elements, including the tabletop, have randomized texture, size, and position. 
\end{itemize}

\subsection{Implementation Details}
\label{appendix:training_details}
\noindent\textbf{Architecture.} We use DINOv2 ViT-L as the image encoder, and a four-layer multilayer perceptron with GELU activations and a final LayerNorm as the robot state encoder.
A transformer backbone comprising 18 alternating global and frame-wise attention blocks~\citep{vggt} is employed to process image and robot state features. 
We initialize the weights of DINOv2 and the transformer backbone from $\pi^3$~\citep{pi3} in order to leverage its generalizable 3D perception capabilities across diverse indoor and outdoor environments.

\noindent\textbf{Training.}
We employ a dynamic number of viewpoints and dynamic batch size during training. Specifically, the number of viewpoints is randomly selected to be either one or two, and the batch size is dynamically set to ensure that each batch contains the same number of images. During training and inference, we use high-resolution images of $630 \times 476$ pixels to capture rich scene details. Data augmentation techniques, including random cropping, color jittering, and Gaussian blur, are applied to the RGB images to enhance data diversity and improve the model's generalization ability. The DINOv2 encoder is kept frozen throughout the training process. \ours is trained on 32 NVIDIA RTX 4090 GPUs for approximately six days. 
The training is performed with a mini-batch size of $384$ images. To reduce GPU memory usage and accelerate training, we employ automatic mixed precision (AMP) with both \texttt{fp32} and \texttt{bf16} data types.
We utilize the AdamW optimizer with a weight decay of $0.05$ and $\mathrm{betas}$ of $(0.9,~0.95)$. The maximum learning rate is set to $2 \times 10^{-5}$. 
The learning rate schedule follows a cosine annealing strategy with linear warmup. Specifically, the learning rate is linearly increased from zero to its maximum value over the first $3{,}000$ iterations. After warmup, it decays following a cosine curve to a minimal value over the remaining training steps.

\subsection{Model Size and Inference Speed}
\label{appendix:model_size_and_inference_speed}
The total number of parameters in \ours is 969.5M.
\ours achieves an inference speed of 43.5 Hz on an NVIDIA RTX 4090 GPU with monocular input, and 18.7 Hz with binocular input, meeting the real-time requirements of robotic manipulation.

\subsection{Real-World Manipulation Experiment Details}
\label{appendix:real-world_experiment_details}

\begin{figure}[t!]
    \centering
    \includegraphics[width=0.98\linewidth]{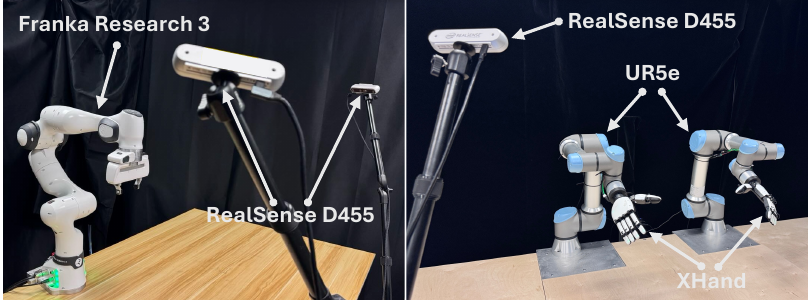}
    \caption{\textbf{Hardware setup.} Our experimental platform features both single-arm and dual-arm robots. We use RealSense D455 cameras to capture images of the scene.}
    \label{fig:real_setup}
    \vspace{-3mm}
\end{figure}

The robotic platforms used in our experiments consist of a single-arm Franka Research 3 equipped with a parallel gripper and a bimanual robot composed of two UR5e arms, each fitted with a 12-DoF XHand. RealSense D455 cameras are utilized to capture both RGB and depth images of the scene. The hardware setup is depicted in Fig.~\ref{fig:real_setup}.

\subsubsection{Imitation Learning Details}
We select four tasks, Sweep Bean, Insert Screw, Breakfast, and BiDex Pour, to validate the application of \ours in imitation learning. Sweep Bean, Insert Screw, and Breakfast are conducted on a single-arm platform, while BiDex Pour is performed on a bimanual robot platform. For the single-arm platform, we use two RealSense D455 cameras to capture images of the scene as input to \ours or the policy, and employ the relative end-effector pose as well as a binary gripper action (open or close) as the action space. For the bimanual robot platform, we use a single camera, and the action space comprises the joint positions of the robot arms and dexterous hands. For the Sweep Bean task, we collect 20 demonstrations, with the broom positioned randomly within a $5 \times 10$ cm region and beans scattered randomly within a $20 \times 20$ cm area. For the Insert Screw task, we collect 80 demonstrations, with the hole randomly positioned within a $15 \times 15$ cm area. For the Breakfast task, we collect 50 demonstrations, with all objects involved randomly placed within separate $10 \times 10$ cm regions. For the BiDex Pour task, we collect 50 demonstrations, with the cup and kettle randomly placed in separate $15 \times 10$ cm areas.

\subsubsection{Sim-to-Real Transfer Details}
\begin{figure}[t!]
    \centering
    \includegraphics[width=0.97\linewidth]{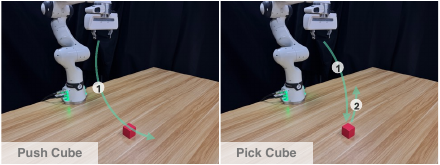}
    \caption{\textbf{Sim-to-real tasks.} We select two tasks for sim-to-real evaluation: Push Cube and Pick Cube.}
    \label{fig:sim2real_tasks}
    \vspace{-3mm}
\end{figure}

\begin{figure}[t!]
    \centering
    \includegraphics[width=0.97\linewidth]{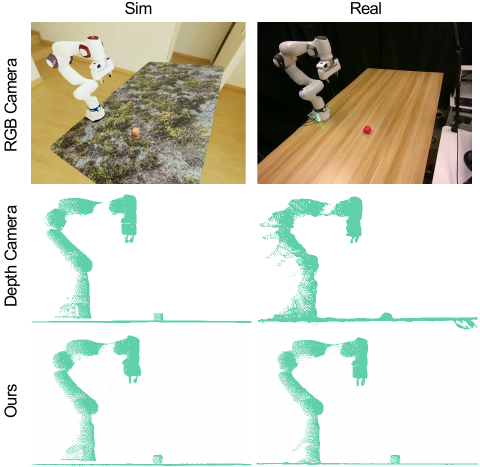}
    \caption{\textbf{Sim-to-real visual gap.} The left column shows data obtained from simulation, while the right column presents data from the real world. The first row displays observations from the RGB camera, the second row shows point clouds from the depth camera, and the third row illustrates point clouds reconstructed by \ours from RGB images.}
    \label{fig:sim2real_visual_gap}
    \vspace{-2mm}
\end{figure}

We assess whether \ours is effective in narrowing the sim-to-real visual gap in two tasks: Push Cube and Pick Cube, as depicted in Fig.~\ref{fig:sim2real_tasks}. The sim-to-real visual gaps produced by different methods are visualized in Fig.~\ref{fig:sim2real_visual_gap}. Compared to RGB images and point clouds acquired by depth cameras, \ours achieves a substantially smaller visual gap.



\subsection{Reconstruction Results in Diverse Scenarios}
\label{appendix:reconstruction_results_in_diverse_scenarios}
\begin{figure}[t!]
    \centering
    \includegraphics[width=0.98\linewidth]{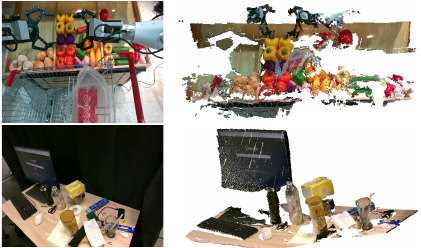}
    \caption{\textbf{Reconstruction results in diverse scenarios.} }
    \label{fig:recon_in_the_wild}
    \vspace{-3mm}
\end{figure}

We evaluate the reconstruction performance of \ours across a variety of scenarios, as illustrated in Fig.~\ref{fig:recon_in_the_wild}. Although \ours is trained on manipulation scenarios involving the Franka robot or the bimanual UR5e robots, it demonstrates strong generalization capabilities to other robot types, such as a wheeled dual-arm robot (Row 1), as well as to indoor scenes without any robots present (Row 2).

\end{document}